# Physics-Informed Machine Learning for Efficient Reconfigurable Intelligent Surface Design

Zhen Zhang[1,2], Jun Hui Qiu[1], Jun Wei Zhang[2], Hui Dong Li[2,*], Dong Tang[1,*], Qiang Cheng[2,*], Wei Lin[3]

[1] *School of Electronics and Communication Engineering, Guangzhou University, Guangzhou 510006, China*
[2] *State Key Laboratory of Millimeter Waves, School of Information Science and Engineering, Southeast University, Nanjing 210096, China*
[3] *Department of Electrical and Electronic Engineering, Hong Kong Polytechnic University, Hong Kong SAR, China*
†Equally contributed to this work.
*Email:
huidongli@seu.edu.cn
tangdong@gzhu.edu.cn
qiangcheng@seu.edu.cn

Zhen Zhang, Junhui Qiu, Dong Tang
School of Electronic and Communication Engineering, Guangzhou University, Guangzhou, 51006, China
Email: zhangzhen@gzhu.edu.cn, qiujunhui@e.gzhu.edu.cn, tangdong@gzhu.edu.cn

Zhen Zhang, Junwei Zhang, Qiang Cheng, Hui-Dong Li
State Key Laboratory of Millimeter Waves, School of Information Science and Engineering, Southeast University, Nanjing 210096, China

Wei Lin
Department of Electrical and Electronic Engineering, The Hong Kong Polytechnic University, Hong Kong SAR, China
Email: w.lin@polyu.edu.hk

**Keywords**: Reconfigurable intelligent surface, design method, data-physics-driven, artificial neural network

**Abstract**
Reconfigurable intelligent surface (RIS) is a two-dimensional periodic structure integrated with a large number of reflective elements, which can manipulate electromagnetic waves in a digital way, offering great potentials for wireless communication and radar detection applications. However, conventional RIS designs highly rely on extensive full-wave EM simulations that are extremely time-consuming. To address this challenge, we propose a machine-learning-assisted approach for efficient RIS design. An accurate and fast model to predict the reflection coefficient of RIS element is developed by combining a multi-layer perceptron neural network (MLP) and a dual-port network, which can significantly reduce tedious EM simulations in the network training. A RIS has been practically designed based on the proposed method. To verify the proposed method, the RIS has also been fabricated and measured. The experimental results are in good agreement with the simulation results, which validates the efficacy of the proposed method in RIS design.

# 1. Introduction

Reconfigurable intelligent surface (RIS) has attracted significant attentions due to its unique abilities to manipulate electromagnetic (EM) waves in recent years. They can control the propagation properties of EM waves dynamically, to enhance the performances of wireless communication systems.[1–3] As shown in **Figure 1**, a RIS usually consists of numerous periodic identical reflective or transmissive elements, and each of them is made of a passive structure and some tunable diodes. By changing the working states of diodes, the current of the RIS element can be altered as well, leading to totally distinct reflection responses to the incident waves.[4-7]

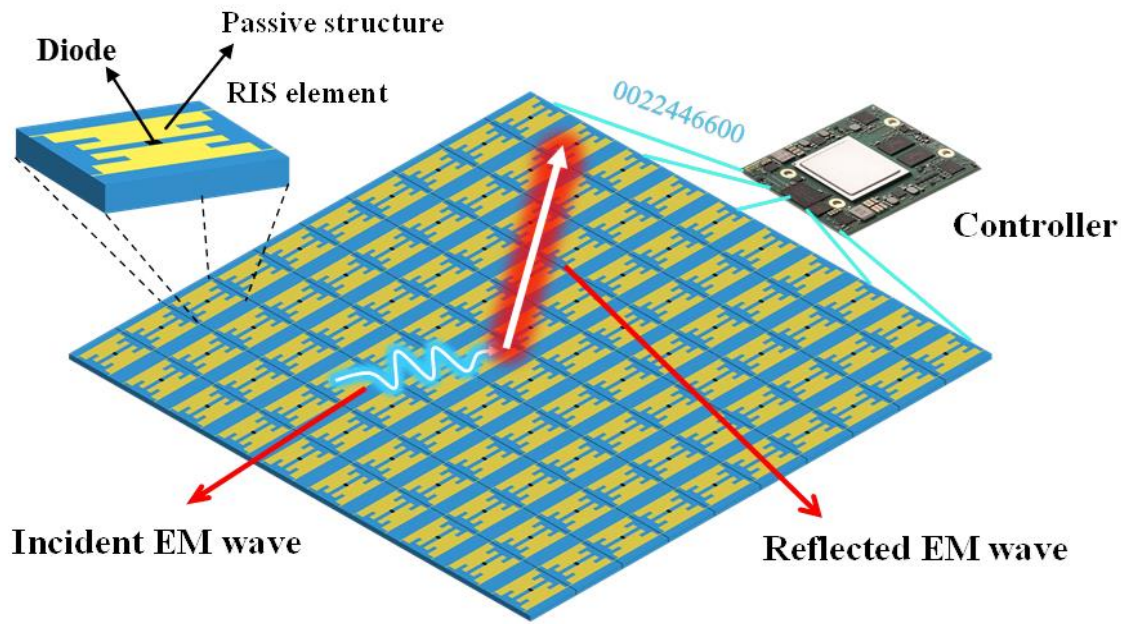

**Figure 1.** Illustrative diagram of a RIS. By tuning the working states of diodes on the RIS element, its reflection coefficients can be effectively altered in real time.

Nowadays, the rapid design of RIS always presents a challenging task since it involves a plenty of design variables in each element, including the dielectric constant of the substrate, the structural parameters of the metallic pattern, and the operational states of diodes.[8,9] At the same time, multiple design objectives have to be fulfilled simultaneously, such as the reflection/transmission amplitude, phase, polarization, and operating bandwidth. To date, three methods have been widely employed to realize the rapid RIS design: the first one is based on expert knowledge that strongly relies on the well-established design schemes reported in the literatures.[10-12] The second method mainly focuses on equivalent circuit models of RIS structures, and then optimizing the structural parameters based on the relationship between geometric and equivalent circuit models.[13-16] The last one relies on structural search algorithms that continuously search and optimize the structural parameters so as to meet the desired EM responses. [17-19] However, the aforesaid three methods are primarily suitable for some typical designs and make delicate adjustments to the structures. When dealing with the design of complex structures, core problems including inefficiency and insufficient model precision still linger.

To address the issue pointed above, the machine-learning-based design method is widely investigated, and is presents an efficient strategy for achieving fast and accurate RIS design. It employs a low-computational-cost machine-learning model to replace the high-computational-cost full-wave EM simulation model, aiming to achieve efficient design and shorten the RIS design period.[20-24] To further reduce the time cost of the full-wave EM simulations, a prior physical knowledge-assisted machine-learning method is proposed,[25–28] which can improve the accuracy and reliability of the model by incorporating existing physical knowledge, and help to reduce the amount of data required for training. However, it still remains a challenge in building accurate physical models to accelerate the modeling process and enhance the design accuracy in machine-learning-assisted RIS design.[29]

To address this problem, we propose a novel physics-driven machine-learning method based on the dual-port microwave network theory, which can precisely characterize the electromagnetic reflective characteristics of the RIS element,[30, 31] and hence significantly accelerate the training of neural networks with reduced computational costs.[32] Specifically, in this work, an optimization framework is constructed to determine the element parameters like the structural dimensions, substrate permittivities and diode models, in order to meet the design goals. Utilizing the proposed method, a 3-bit RIS has been designed, fabricated and measured. The experimental results are in excellent agreement with the theoretical ones, verifying the effectiveness of the proposed method.

## 2. Analysis of Design Theory for RIS

Each RIS element is made of two parts: the passive metallic pattern and the tunable diodes. Correspondingly, it is essential for us to carefully design topological forms, refine structural parameters, and choose the optimal commercial diodes. These principles aim to attain ideal traits like minimal reflection loss, an extensively large phase tuning range, and a wide operation bandwidth. To quickly evaluate the influence of the diodes on the reflection coefficient of the RIS element, a dual-port network is introduced to describe the reflective properties of the RIS element.[30] As shown in **Figure 2**, one port (Port 1) is used to monitor the incident/reflected EM waves, and the other port (Port 2) is an internal port to connect the diode with the passive structures. In this way, the characteristics of passive structures and diodes can be considered separately within the microwave network, significantly enhancing the simulation efficiency of the RIS element. Specifically, when the passive structures or diodes change, there is no need of complex field-circuit co-simulation. And it is only necessary to modify the EM responses of the relevant structures or devices within the network.

**Figure 2b** gives the dual-port network model of the RIS element, where $\mathbf{Z}$ is the impedance matrix of the passive structure as follows:

$$\mathbf{Z}(\omega) = \begin{bmatrix} Z_{11}(\omega) & Z_{12}(\omega) \\ Z_{21}(\omega) & Z_{22}(\omega) \end{bmatrix}, \quad (1)$$

where $\omega$ represents the operating angular frequency. $Z_{11}$ and $Z_{22}$ denote the self-impedances at port 1 and 2 respectively. $Z_{12}$ and $Z_{21}$ are the mutual impedances between port 1 and port 2. $Z_{12}$ is the transpose of $Z_{21}$, $Z_a$ is the input impedance of the commercial diode. $Z_0$ represents the wave impedance in free space. From the microwave network theory, the reflection coefficient of the RIS element can be calculated by:

$$S_{11}(\omega) = \frac{Z_{11}(\omega) - \dfrac{Z_{12}(\omega) \cdot Z_{21}(\omega)}{Z_a(\omega) + Z_{22}(\omega)} - Z_0(\omega)}{Z_{11}(\omega) - \dfrac{Z_{12}(\omega) \cdot Z_{21}(\omega)}{Z_a(\omega) + Z_{22}(\omega)} + Z_0(\omega)} \quad (2)$$

It is clear that from Equation (2) we can directly calculate the reflection coefficient when the impedance of the diode is changed without any EM simulation. Therefore, it can assist us in quickly screening the commercial diodes that match the passive structure best.

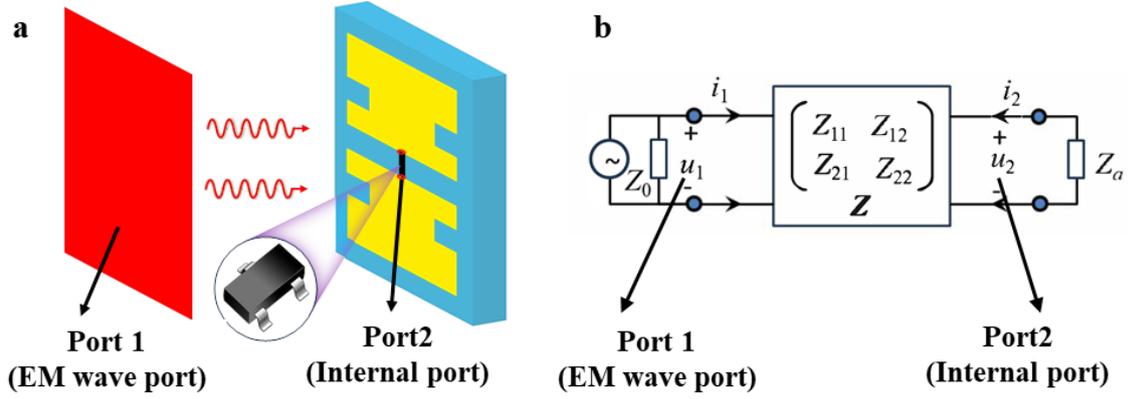

**Figure 2.** (a) Schematic of the RIS element and (b) dual-port network of the RIS element, where the port 1 represents the plane wave excitation, and $Z_a$ at port 2 is the impedance of the diode.

As seen in Equations (1-2), the impedance matrix **Z** is essential to determine the reflection coefficient of the RIS element, which is usually acquired from EM simulations. However, during the process of structural optimization, any change in geometric shape requires the repetition of long-time electromagnetic simulations, which makes the optimization efficiency extremely low. To tackle this challenge, the multilayer perceptron (MLP), a widely-used machine-learning method, is used to establish the relationship between the geometric parameters of the passive structure and the impedance matrix **Z**. The framework of the MLP model consists of three layers: the input layer, the hidden layer, and the output layer.

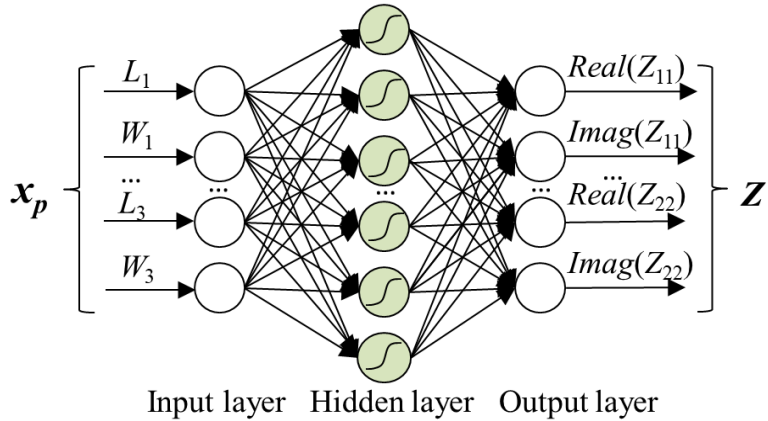

**Figure 3.** Framework of the MLP model. The MLP neural network is a three-layer feedforward neural network, where the input layer contains the passive structural parameters [$L_1$, $W_1$, $L_2$, $W_2$, $L_3$, $W_3$] of the RIS element. The hidden layer employs the Sigmoid function [34]. And the final output layer contains the real and imaginary parts of the impedance matrix **Z**.

The framework of the used three-layer MLP is shown in **Figure 3**. The input layer of the model is constituted by the geometric parameters of the RIS element, while the output layer corresponds to the impedance matrix **Z**. Within the design space, we utilize Monte Carlo sampling method [33] to randomly generate vectors of geometric parameters, as the inputs of the MLP. The hidden layer comprises 30 neurons. In this MLP model, each neuron ingests inputs from the preceding layer of neurons, conducts a weighted summation operation and appends a biasing term. Subsequently, the outcome of this computation is fed into the activation function. This activation function contains the Sigmoid function, which plays a crucial role in transforming inputs into an output value. This output value, in turn, serves as the input for the subsequent layer of neurons, thereby facilitating the sequential flow of data processing and enabling the model to effectively learn and make predictions based on the input-output relationships embedded within its structure. As the impedance is a complex number, both the real and imaginary parts are considered in the output of the MLP model.

Once the training is finished based on the numerical simulations, the MLP model can provide fast and accurate predictions of the impedance matrix $Z$ for passive structure with arbitrary combination of geometric parameters. By incorporating the impedance matrix into the dual-port network, we can rapidly obtain the reflection coefficient $S_{11}$ of the RIS element. The proposed integrated model, built on the MLP (Multi-Layer Perceptron) and DPN (Dual-Port Network), is termed the MLP-DPN model in the remaining part of the paper. Thereafter, by employing the established model, we are capable of promptly computing the $S_{11}$ of the RIS elements with assorted passive structural dimensions and distinct diodes, thus eliminating the necessity for EM simulations. The computational formula of the MLP-DPN model is recapitulated as follows:

$$S_{11}(x_p, x_a) = F(Z_{MLP}(x_p), x_a), \quad (3)$$

where $x_p$ denotes the vector of passive structural parameters of the RIS element. $x_a$ represents the circuit parameters of the diode. $Z_{MLP}(x_p)$ describes the relationship between $x_p$ and the impedance matrix $Z$, and $F(Z_{MLP}(x_p), x_a)$ is the reflection coefficient function of $Z_{MLP}(x_p)$ and $x_a$.

Based on the MLP-DPN model in **Equation** (3), we utilize an optimization algorithm to identify the design parameters that satisfy the design requirements. These parameters encompass both the parameters of passive structure $x_p$ and those of diodes $x_a$. Among them, the former is closely related to the current distribution on the RIS element, and the latter has distinct impedances in different switching states. Both of them contribute to the electromagnetic characteristics of the element simultaneously. During the optimization process, the adjustment of these two types of parameters also needs to be carried out simultaneously. Taking the $N$-bit metasurface with $2^N$ phase states as an example, we define the phase difference $\Delta\theta$, phase state $\theta^{(k)}$, and the matching function $L(x_p, x_a^{(k)}, \emptyset^{(k)})$ in Equations (4-6) respectively as:

$$\Delta\theta = |\varphi(S_{11}(x_p, x_a^{k+1})) - \varphi(S_{11}(x_p, x_a^k))|, \quad (4)$$

$$\theta^{(k)} = (k-1) \cdot \Delta\theta, \quad (5)$$

$$L(x_p, x_a^{(k)}, \emptyset^{(k)}) = \sum_{K=1}^{2^N} w_1 \cdot \left(|\varphi(S_{11}(x_p, x_a^k)) - \theta^{(k)}|^2\right) - w_2 \cdot (\max(|S_{11}(x_p, x_a^k)|_{dB}, A)), \quad (6)$$

where $k$ stands for the number of phase state with $k = 1,\ldots, 2^N$. $\theta^{(k)}$ is the reflection phase at the $k$-th phase state. $\Delta\theta$ is the phase difference between the $k$-th and the $(k+1)$-th phase states. $L$ is utilized to evaluate the sum of the differences of both the reflection phase and amplitude between the obtained values and the expected ones. $w_1$ and $w_2$ are the phase and amplitude matching weights in the $L$ function. Here we choose $w_1 = w_2 = 0.5$. $S_{11}(x_p, x_a^k)$ is the reflection coefficient at the $k$-th state. $|\cdot|_{dB}$ is the dB form of amplitude of the reflection coefficient •. $\varphi$ is the reflection phase, and $A$ is threshold of the minimum reflection amplitude. $\|\cdot\|_2$ is the 2-norm, and max( • , △) is the operation to find the maximum value between • and △. $w_1$ and $w_2$ are the proportions of phase matching and amplitude constraints in the objective function respectively. $\emptyset^{(k)} = [\theta^{(1)}, \theta^{(2)}, \ldots, \theta^{(k)}]$ is the phase state vector, where $k = 1,\ldots, 2^N$.

For example, when $k = 1$, we select the initial parameter of the diode $x_a^{(1)}$ to calculate the first phase state with $\emptyset^{(1)} = \theta^{(1)}$. Then we proceed to perform $2^N$ rounds of iteration. The details of the optimization algorithm in the *k*-th iteration contains the following three steps:
1) Optimize the passive structure parameters to get $x_p^{(k+1)}$ when $x_a^{(k)}$ $x_p^k$ and $\emptyset^{(k)}$ are given,
$$x_p^{(k+1)} = \arg\min_{x_p} L(x_p, x_a^{(k)}, \emptyset^{(k)}), \tag{7}$$
2) Update $x_a^{(k+1)}$ when $x_p^{(k+1)}$ and $x_a^{(k)}$ are determined:
$$x_a^{(k+1)} = \arg\min_{x_a} L(x_p^{(k+1)}, x_a, \emptyset^{(k)}), \tag{8}$$
3) Update $\emptyset$:
$$\emptyset^{(k+1)} = [\emptyset^{(k)}, \theta^{(k+1)}]. \tag{9}$$
After $2^N$ iterations, we can get the desired RIS element with $2^N$ phase states.

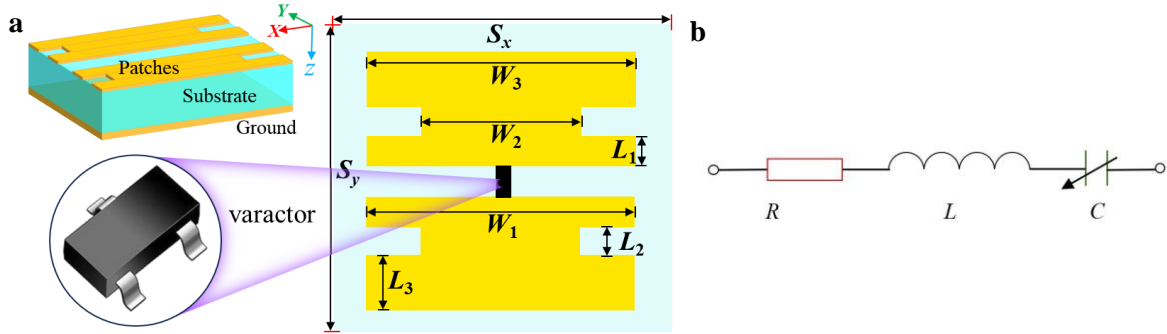

Figure 4. (a) Configuration of the RIS element. (b) Equivalent circuit of the diode.

## 3. Simulation Results

In this section, the performance of the proposed method is validated through the design and implementation of a 3-bit RIS element with eight reflection phase states, specifically 0°, 45°, 90°, 135°, 180°, 225°, 270°, and 315°. And the amplitude of the reflection coefficient at the working frequency is always greater than –3dB. The numerical simulations are executed using the EM solver, namely CST Microwave Studio, on a computer system equipped with an AMD Ryzen 5500 CPU and 32 GB of random access memory (RAM).

An initial RIS element is inspired by the structure in [11], which is illustrated in **Figure 4(a)**. The element is implemented on a F4B substrate with $\varepsilon_r = 2.65$, $\tan\delta = 0.001$ and the thickness is 3.3 mm. Two metallic patches are bridged by a varactor diode (SMV-2019-079LF) on the top of the element, with the equivalent circuit shown in **Figure 4(b)**. Here $R = 0.3\,\Omega$, $L = 0.7$nH, and $C$ is the equivalent capacitance ranging from 0.6 pF to 2.6 pF depending on the biasing voltage. [$W_1$, $W_2$, $W_3$] and [$L_1$, $L_2$, $L_3$] stand for the widths and lengths of the metal patches on the top layer. The vector of the design variable for the passive structure is $x_p = [W_1, W_2, W_3, L_1, L_2, L_3]$. Since only one varactor diode is employed in the element design, the design variable of the diode is the equivalent capacitance $x_a = C$. The upper and lower bounds of the design parameters for the RIS element are listed in **Table 1**.

Table 1. The Upper and Lower Bounds of the Design Parameters for the RIS Element

| | Passive structure | | | | | | Varactor |
|---|---|---|---|---|---|---|---|
| Variables (mm) | $W_1$ | $W_2$ | $W_3$ | $L_1$ | $L_2$ | $L_3$ | $C$ (pF) |
| Maximum | 6 | 6 | 6 | 1.0 | 0.1 | 6.0 | 0.6 |
| Minimum | 24 | 24 | 24 | 3.0 | 1.0 | 8.0 | 2.6 |

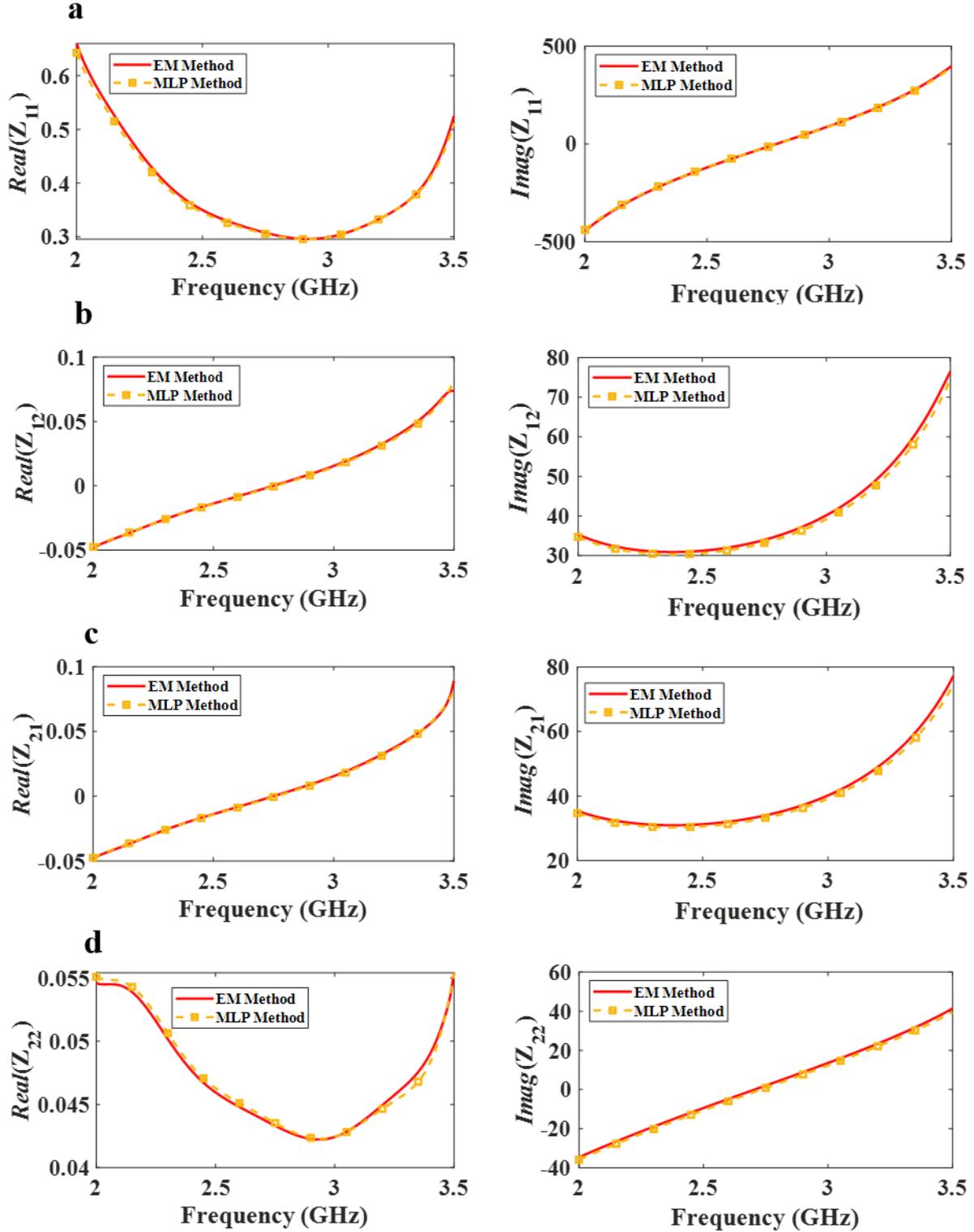

**Figure 5.** Modeling results comparison of the 3-bit RIS element between the MLP model and the full-wave EM model: real and imaginary parts of (a) $Z_{11}$, (b) $Z_{12}$, (c) $Z_{21}$, and (d) $Z_{22}$.

The MLP model's accuracy is crucial for evaluating the proposed design method's performance. Within the design step of the passive structure, 2000 samples are randomly sampled using the Monte Carlo sampling method, with 80% are training samples and 20% are testing samples. Each electromagnetic (EM) simulation takes 36 seconds, and the whole process lasts for 20 hours in total. In this example, the inputs of the MLP model are $x_p$ and $x_a$, while the outputs are $Real(Z_{11})$, $Imag(Z_{11})$, $Real(Z_{12})$, $Imag(Z_{12})$, $Real(Z_{21})$, $Imag(Z_{21})$, $Real(Z_{22})$, $Imag(Z_{22})$, in which *Real* and *Imag* represent the real and imaginary part of the variable respectively. The entire process of building the MLP model costs 0.15 hours. To test the performance of the MLP model, we calculate the mean square error (MSE) and mean absolute error (MAE) between the impedances obtained from the EM simulation and the MLP model, as shown in **Table 2**. The detailed metrics are defined as follows:

$$MSE = \frac{1}{K}\sum_{k=1}^{K}\left|Z_{i,j}^{k} - \hat{Z}_{i,j}^{k}\right|^{2} \quad (4)$$

$$MAE = \frac{1}{K}\sum_{k=1}^{K}\left|Z_{i,j}^{k} - \hat{Z}_{i,j}^{k}\right| \quad (5)$$

in which *k* is the number of the testing data. *i*, *j* are the waveport number with *i* = 1, 2, and *j* = 1, 2. $\widehat{Z_{i,j}^{k}}$ is the predicted value of the *k*-th sample from the MLP model, and $Z_{i,j}^{k}$ is the real or imaginary value of the *k*-th sample from EM simulation. **Figure 5** also shows the predicted **Z** matrix components with the Multi-Layer Perceptron (MLP), which match well with the full-wave simulation results, demonstrating the effectiveness of the training network.

Table 2. MSE and MAE of the MLP Model for the RIS Element

| Errors | Training | | Testing | |
|---|---|---|---|---|
| | MSE | MAE | MSE | MAE |
| $Real(Z_{11})$ | 0.00056 | 0.00043 | 0.00065 | 0.00049 |
| $Real(Z_{12})$ | 0.00015 | 0.00009 | 0.00015 | 0.00009 |
| $Real(Z_{21})$ | 0.00015 | 0.00009 | 0.00015 | 0.00010 |
| $Real(Z_{22})$ | 0.00034 | 0.00021 | 0.00034 | 0.00025 |
| $Imag(Z_{11})$ | 0.07446 | 0.05579 | 0.09136 | 0.06923 |
| $Imag(Z_{12})$ | 0.04526 | 0.03607 | 0.04747 | 0.03880 |
| $Imag(Z_{21})$ | 0.04842 | 0.03763 | 0.05082 | 0.03994 |
| $Imag(Z_{22})$ | 0.07065 | 0.01716 | 0.02301 | 0.02216 |

**Table 2** displays the corresponding MSE and MAE of the MLP model for the RIS element. It can be found that for both the test set and the training set, the mean squared error (MSE) of the real and imaginary parts of the elements in **Z** matrix is always lower than 0.1, and the mean absolute error (MAE) is less than 0.07. This indicates that the adopted multi-layer perceptron (MLP) model has high accuracy and stability when calculating the **Z** impedance for different passive structures.

The constructed MLP model is combined with the aforementioned DPN model to rapidly determine the reflection coefficients of the RIS element with different passive structures and different diodes, which can further combine with the optimization algorithms to obtain the optimal element to meet the design requirements. To further validate the MLP-DPN model, three RIS elements (Element 1-3 in **Figure 6 (a)**) with center frequencies of 2.68 GHz, 3.14 GHz, and 3.3 GHz are designed based on this model, as seen in **Figure 6**. The corresponding structural and diode parameters are listed in **Table 3**. The reflection amplitude and phase spectra of the three 3bit elements are illustrated in **Figures 6 (b), (c) and (d),** respectively. The reflection phase errors of the three elements at the corresponding frequencies are below 5 °, and the reflection amplitudes are greater than -3 dB. The design process of each RIS element takes a very short time less than 0.2 hour.

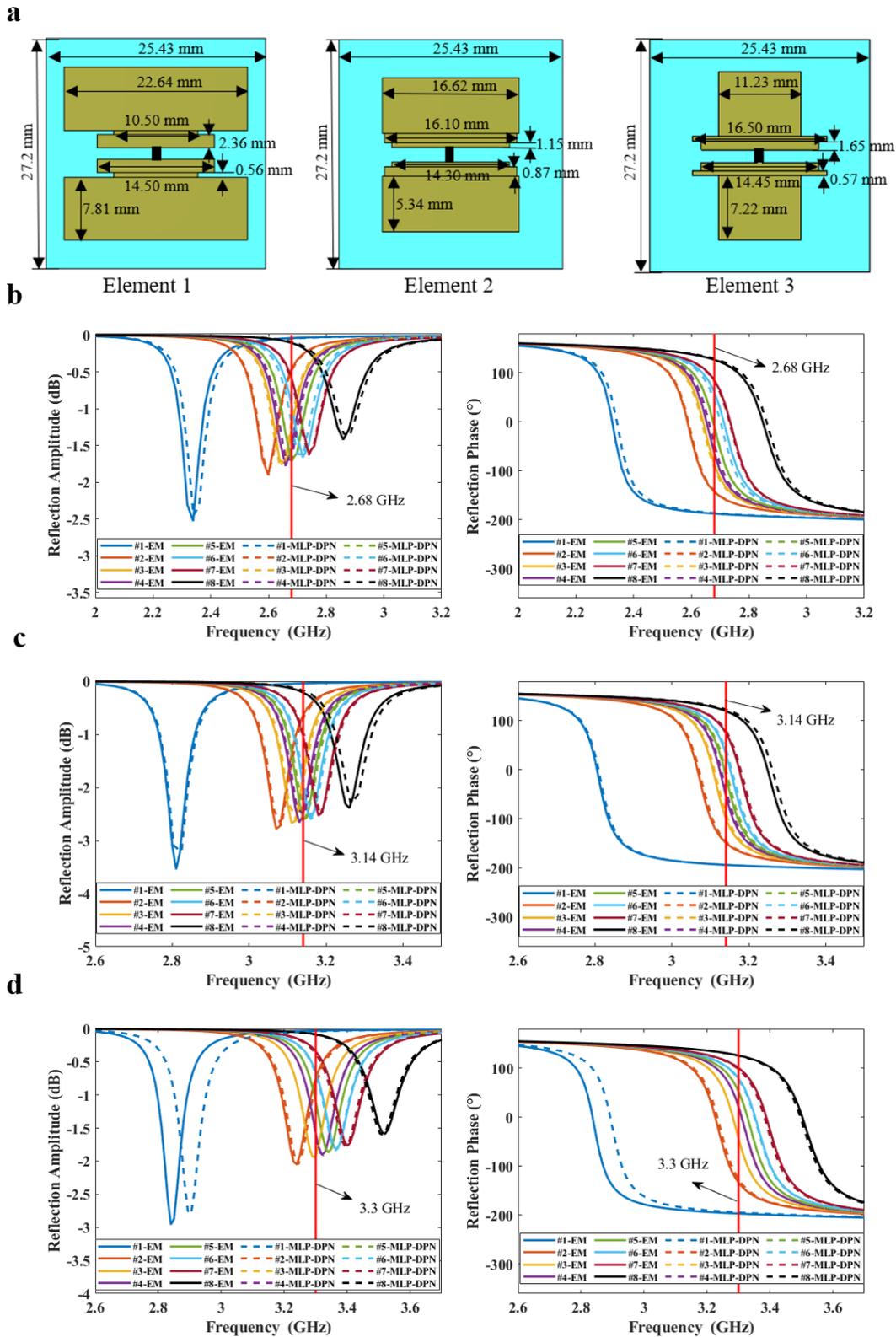

**Figure 6.** (a) Cofigurations of three RIS elements, which operate at frequencies of 2.65 GHz, 3.15 GHz, and 3.3 GHz, respectively. Reflection amplitude and phase of (b) Element 1, (c) Element 2 and (d) Element 3. In (b), (c) and (d), the solid lines represent the results of the electromagnetic (EM) simulation, while the dashed lines represent the results obtained by utilizing the MLP-DPN model. 1-EM, 2-EM, …, 8-EM denote the 3-bit working states of the EM simulation, and 1-MLP-DPN, 2-MLP-DPN, …, 8-MLP-DPN denote the 3-bit working states based on the MLP-DPN model.

It is worthy noting that, in the traditional design method, numerical simulation is required for every change of the structural parameters. Taking the structure shown in **Figure 6 (a)** as an example, suppose there are six structural parameters for each element and each parameter has six possible values. In this case, a total of $6^6$ EM simulations would be required. Meanwhile, the changes of the equivalent circuits for the diode caused by the variation of biasing voltage also demand an increase of EM simulations. This process is not only time-consuming and laborious, but also places a high burden on the computing resources. In this work, as a comparison, the MLP-DPN model method only needs EM simulations in the initial stage of constructing the MLP model. In the subsequent design of multiple RIS elements, there is no need for EM simulation at all, which significantly improves the design efficiency. To illustrate this point, we provide the time comparison between the traditional and the proposed methods in the designs of the above three elements, as shown in **Table 4**. It can be seen that the time cost of our method is only 3.7% of the counterpart of the traditional method. Moreover, as the number of RIS elements to be designed increases, the advantages become even more pronounced.

**Table 3.** Structural and Diode Parameters of Three RIS Elements in Figure 6

| Operation Frequency /GHz | Optimization of passive structural parameters | Optimization of tunnable device parameters | | | | | | | | |
|---|---|---|---|---|---|---|---|---|---|---|
| | | Different working states | 1 | 2 | 3 | 4 | 5 | 6 | 7 | 8 |
| 2.68 | $x_p^*$ /mm [14.50, 10.50, 22.64, 2.36, 0.56, 7.81] | $x_a^*$ /pf | 2.60 | 1.45 | 1.26 | 1.22 | 1.21 | 1.15 | 1.10 | 0.95 |
| | | Phase /° | 0 | 44.9 | 91.0 | 139.4 | 178.3 | 230.0 | 268.6 | 315.8 |
| | | Amplitude /dB | -0.05 | -0.48 | -1.17 | -1.69 | -1.72 | -1.20 | -0.62 | -0.09 |
| 3.14 | [14.30, 16.10, 16.62, 1.15, 0.87, 5.34] | $x_a^*$ /pf | 2.60 | 1.37 | 1.18 | 1.15 | 1.13 | 1.11 | 1.07 | 0.97 |
| | | Phase /° | 0 | 44.7 | 94.2 | 140.7 | 180.6 | 226.1 | 268.4 | 315.2 |
| | | Amplitude /dB | -0.04 | -0.64 | -1.74 | -2.52 | -2.61 | -1.99 | -1.04 | -0.19 |
| 3.3 | [14.45, 16.50, 11.23, 1.65, 0.57, 7.22] | $x_a^*$ /pf | 2.60 | 1.13 | 1.01 | 0.99 | 0.95 | 0.92 | 0.88 | 0.77 |
| | | Phase /° | 0 | 44.5 | 88.1 | 138.4 | 182.4 | 228.5 | 269.2 | 315.0 |
| | | Amplitude /dB | -0.03 | -0.47 | -1.16 | -1.79 | -1.84 | -1.37 | -0.71 | -0.13 |

**Table 4.** Comparison Between the Traditional and the Proposed Methods for Three RIS Elements

| Design method | Time cost /hour | | | Total time /hour |
|---|---|---|---|---|
| | | Modeling | Design | |
| Our proposed method | Element 1 | 20.15 | 0.2 | 20.75 |
| | Element 2 | -- | 0.2 | |
| | Element 3 | -- | 0.2 | |
| Traditional method | Element 1 | 207 | | 559 |
| | Element 2 | 169 | | |
| | Element 3 | 183 | | |

-- refers to increase unnecessary time costs. Element 1, Element 2, and Element 3 operate at 2.68GHz, 3.14GHz and 3.3GHz.

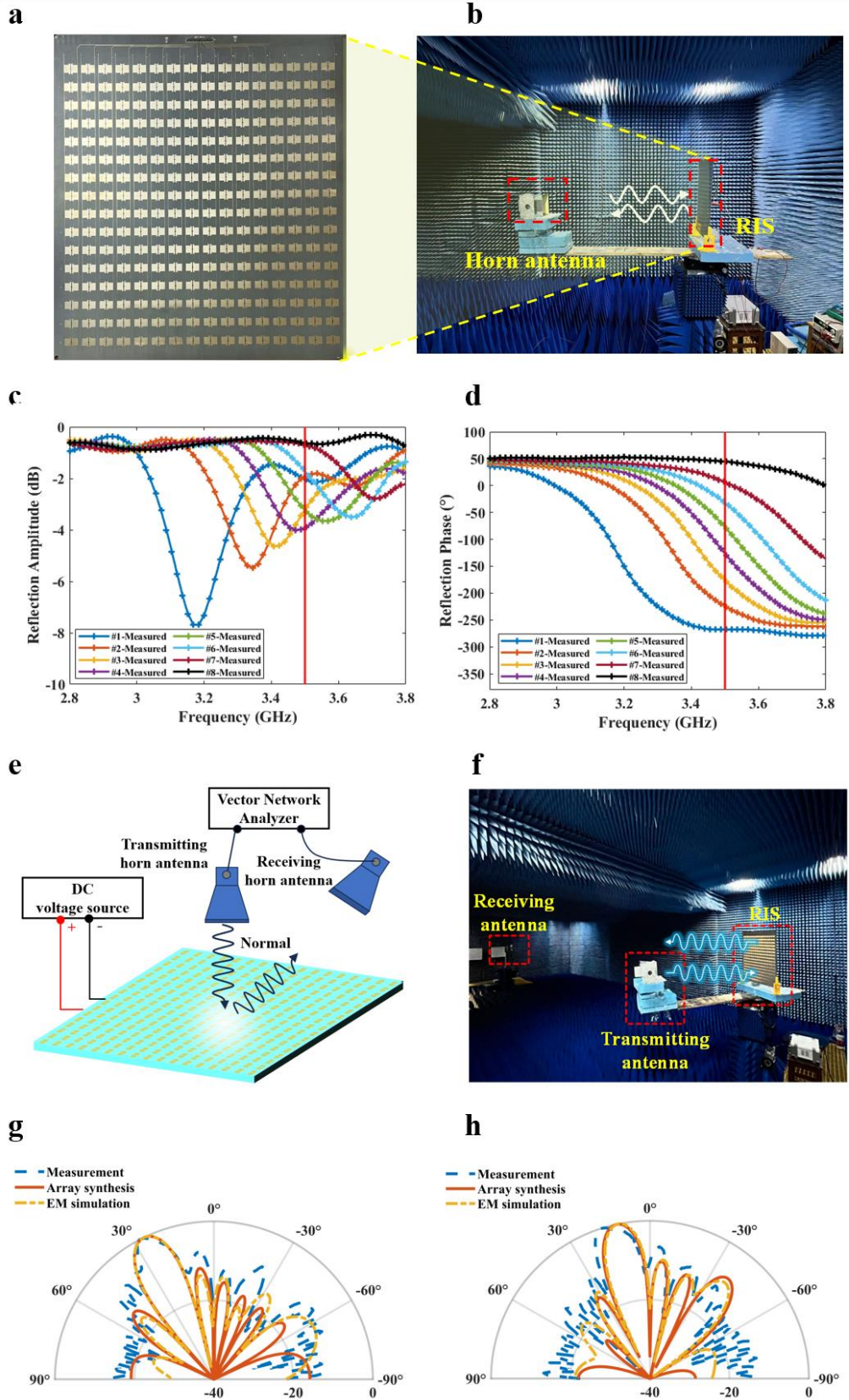

**Figure 7.** (a) Photograph of the fabricated 16×16 RIS. (b) Measurement setup of RIS for EM reflection coefficient. Measured reflection (c) amplitudes and (d) phases of RIS under 8 states. (e-f) Measurement setup of RIS for far-field scattering pattern. Far-field scattering patterns with (g) "1111333355557777" and (h) "1133557711335577" phase codes.

## 4. Measurement results

To further validate this method, we have fabricated a 3bit RIS based on Element 3 (see **Figure 6 (a)**) and made the corresponding measurements. The photograph of the fabricated prototype is shown in **Figure 7 (a),** which consists of 16 ×16 elements in all. The total size of the RIS is 436 mm ×490 mm. There are 256 varactor diodes mounted on the sample. All the varactors in the same column share the same biasing voltage, indicating all the elements in each column can be tuned simultaneously.

The reflection coefficient of the sample is measured from 2.8 GHz – 3.8 GHz under different biasing voltages in the microwave anechoic chamber. A horn antenna is connected to the VNA (Agilent N5245A) to transmit and receive EM waves. The center of the antenna aligns with the center of the metasurface in height. A DC voltage source is employed to supply the required biasing voltage to the diodes in different columns. A metal plate with the same size as the RIS is also measured for calibration. The measured reflection amplitude and phase spectra are presented in **Figures 7 (c-d)** respectively. It can be observed that at 3.5 GHz, the RIS has eight phase states: 0 °, 43.8 °, 91.6 °, 141.5 °, 190.7 °, 235.1 °, 274.1 ° and 313.0 °, with the magnitude of all eight states' reflection coefficients being less than –3.93 dB. Compared to the simulation results, there is a 0.2 GHz frequency offset in the test results. This is primarily caused by the processing inaccuracies, poor soldering, and variations in control voltage.

The far field scattering pattern is also measured at 3.5GHz in a microwave anechoic chamber. Here two horn antennas are employed for transmitting and receiving signals, respectively. Sixteen independent DC voltage sources are employed to precisely control the columns of RIS elements. **Figure 7 (e-f)** show the schematic diagram of the experiment setup and the test environment. The transmitting antenna and the RIS are mounted on a turntable, while the receiving horn antenna is positioned 8 meters away from the RIS. To effectively suppress unwanted reflections from the surrounding environment, time-gating technology is implemented in the experimental setup.

In the experiment, the scattering pattern is varied by adjusting the reflection phase of each column. The coding state of each column in the RIS array is determined by the desired beam direction. From the generalized Snell's law, we can see that when the beam angle is set to 30 °, the corresponding coding sequence "1111333355557777", while as for the angle of 15 °, the coding sequence is "1133557711335577". Here we use the digits 0-7 to represent the eight phase states of 3bit RIS from 0 °, 45 °, 90 °,135 °, 180 °, 225 °, 275 °, 315 °. The calculated and measured scattering patterns in the two cases are plotted in **Figure 7(g-h)** respectively. There are slight differences in the sidelobes, which are mainly attributed to the mutual coupling among adjacent RIS elements in the array synthesis method [30]. However, the predicted and measured scattering patterns show reasonable agreement with each other, which validates the presented method.

## 5. Conclusions

In this work, we propose a fast reconfigurable intelligent surface (RIS) design method based on the meachine-learning and microwave network theory. The proposed hybrid method contains three key factors: the Multi-Layer Perceptron (MLP), the dual-port network (DPN), and the optimization algorithm. Specifically, the reflection coefficient of RIS element is obtained from the MLP-DPN model, aiming to circumvent some time-consuming EM simulations and such develop a cost-effective surrogate model. After that, based on the MLP-DPN model, a optimization algorithm is implemented for parameter optimization of the RIS element. Finally, to validate the performance of this proposed method, we have designed three 3-bit reflective phase-modulated RISs, one of which was fabricated and measured. The obtained reflection coefficients and far-field scattering patterns demonstrate that our proposed design method is effective for fast and accurate RIS design.

## Acknowledgments

This work is supported by the National Key Research and Development Program of China (grant no.2023YFB3811502); the National Science Foundation (NSFC) for Distinguished Young Scholars of China (grant no.62225108), the National Natural Science Foundation of China (grant no.62288101, grant no.62201139, grant no.U22A2001), the Jiangsu Province Frontier Leading Technology Basic Research Project (grant no.BK20212002), the Jiangsu Provincial Scientific Research Center of Applied Mathematics (grant no.BK20233002), the Program of Song Shan Laboratory (Included in the management of Major Science and Technology Program of Henan Province) (grant no.221100211300-02, grant no.221100211300-03), the 111 Project (111-2-05), the Fundamental Research Funds for the Central Universities (grant no.2242022k60003, grant no.2242024RCB0005), the Southeast University - China Mobile Research Institute Joint Innovation Center (grant no.R202111101112JZC02), and the State Key Laboratory of Millimeter Waves, Southeast University (grant no.K202403), and the Basic Innovation Program for Post Graduate Students by Guangzhou University under Grant (JCCX2024-037).